\documentclass[11pt,a4paper]{article}
\usepackage[hyperref]{emnlp2020}
\usepackage{times}
\usepackage{latexsym}

\usepackage{amsmath}
\usepackage{amsthm}
\usepackage{amsfonts}
\usepackage{amssymb}
\usepackage{xspace}
\usepackage{xfrac}
\usepackage{csquotes}
\usepackage{booktabs}
\usepackage[T5,T1]{fontenc}
\usepackage{combelow}

\DeclareTextSymbolDefault{\ohorn}{T5}
\DeclareTextSymbolDefault{\uhorn}{T5}

\usepackage{tipa}
\usepackage{graphicx} 
\graphicspath{{./figs/}} 

\newcommand{\vh}{\mathbf{h}}
\newcommand{\vW}{\mathbf{W}}
\newcommand{\vM}{\mathbf{M}}
\newcommand{\vV}{\mathbf{V}}
\newcommand{\vU}{\mathbf{U}}

\newcommand{\calT}{\mathcal{T}}
\newcommand{\R}{\mathbb{R}}
\newcommand{\softmax}{\mathrm{softmax}}
\newcommand{\defn}[1]{\textbf{#1}}

\newcommand{\rank}{\mathrm{rank}}

\newcommand{\bert}{BERT\xspace} 
\newcommand{\albert}{ALBERT\xspace} 
\newcommand{\roberta}{RoBERTa\xspace} 
\newcommand{\elmo}{ELMo\xspace}
\newcommand{\fasttext}{fastText\xspace}
\newcommand{\tr}{\mathrm{tr}}

\definecolor{C1}{rgb}{0.2823529411764706, 0.47058823529411764, 0.8156862745098039}
\definecolor{C3}{rgb}{0.1, 0.7, 0.1}
\definecolor{C4}{rgb}{0.9, 0, 0}
\definecolor{C5}{HTML}{6A0DAD}
\definecolor{C6}{rgb}{0.5490196078431373, 0.3803921568627451, 0.23529411764705882}
\definecolor{C7}{rgb}{0.8627450980392157, 0.49411764705882355, 0.7529411764705882}
\definecolor{C8}{rgb}{0.4745098039215686, 0.4745098039215686, 0.4745098039215686}

\newcommand{\legendalbert}{{\color{C1} \albert}} 
\newcommand{\legendbert}{{\color{C7} \bert}}
\newcommand{\legendroberta}{{\color{C6} \roberta}} 
\newcommand{\legendfasttext}{{\color{C3} \fasttext}} 
\newcommand{\legendonehot}{{\color{C4} one-hot}} 
\newcommand{\legendrandom}{{\color{C5} random}} 

\usepackage{cleveref}
\crefname{section}{\S}{\S\S}
\Crefname{section}{\S}{\S\S}
\crefname{table}{Tab.}{}
\crefname{figure}{Fig.}{Figs.}
\crefname{algorithm}{Alg.}{}
\crefname{appendix}{App.}{}
\crefname{lemma}{Lemma}{}
\Crefname{theorem}{Theorem}{}
\crefname{prop}{Proposition}{}
\crefname{cor}{Corollary}{}
\crefname{align}{eq.}{}
\crefname{equation}{eq.}{}

\newcommand{\citeposs}[1]{\citeauthor{#1}'s (\citeyear{#1})}

\usepackage{microtype}

\aclfinalcopy %
\def\aclpaperid{1551} %

\everypar{\looseness=-1}

\title{Pareto Probing: Trading Off Accuracy for Complexity}

\usepackage{emoji}
\newcommand{\ucambridge}{\emoji[twitter]{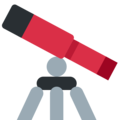}}
\newcommand{\uedinburgh}{\emoji[twitter]{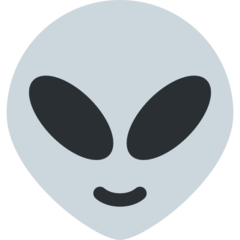}}
\newcommand{\ethz}{\emoji[twitter]{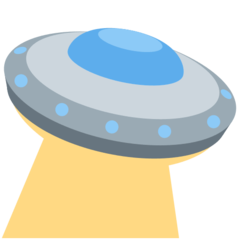}}
\newcommand{\fairesearch}{\emoji[twitter]{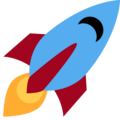}}

\author{
Tiago Pimentel$\thanks{~~Equal contribution}$ $^{\,,}$\raise1.0ex\hbox{\normalfont\ucambridge}~\;~ Naomi Saphra$^{*,}$\raise1.0ex\hbox{\normalfont\uedinburgh}~\;~ Adina Williams\raise1.0ex\hbox{\normalfont\fairesearch}~\;~ Ryan Cotterell\raise1.0ex\hbox{\normalfont\ucambridge\!,\ethz}\\
  \raise1.0ex\hbox{\normalfont\ucambridge}University of Cambridge~\;~\raise1.0ex\hbox{\normalfont\uedinburgh}University of Edinburgh~\;~\raise1.0ex\hbox{\normalfont\fairesearch}Facebook AI Research~\;~\raise1.0ex\hbox{\normalfont\ethz}ETH Z\"{u}rich \\
  \texttt{tp472@cam.ac.uk},~\;~ \texttt{n.saphra@ed.ac.uk}, \\
  \texttt{adinawilliams@fb.com},~\;~ \texttt{ryan.cotterell@inf.ethz.ch}
}

\date{}

\begin{document}
\maketitle
\begin{abstract}
The question of how to probe contextual word representations for linguistic structure in a way that is both principled and useful has seen significant attention recently in the NLP literature.
In our contribution to this discussion, we argue for a probe metric that reflects the fundamental trade-off between probe complexity and performance: the Pareto hypervolume. 
To measure complexity, we present a number of parametric and non-parametric metrics. 
Our experiments using Pareto hypervolume as an evaluation metric show that probes often do not conform to our expectations---e.g., why should the non-contextual \fasttext representations encode more morpho-syntactic information than the contextual \bert representations?
These results suggest that common, simplistic probing tasks, such as part-of-speech labeling and dependency arc labeling, are inadequate to evaluate the linguistic structure encoded in contextual word representations. 
This leads us to propose full dependency parsing as a probing task.
In support of our suggestion that harder probing tasks are necessary, our experiments with dependency parsing reveal a wide gap in syntactic knowledge between contextual and non-contextual representations.
Our code can be found at \url{https://github.com/rycolab/pareto-probing}.
\looseness=-1
\end{abstract}
\section{Introduction}  
Neural networks are a pillar of modern NLP systems.
However, their inner workings are poorly understood; indeed, for this reason, they are often referred to as black-box systems \cite{psichogios1992hybrid,orphanos1999decision,cauer2000life}.
This lack of understanding, coupled with the rising adoption of neural NLP systems in both industry and academia, has fomented a rapidly growing literature devoted to ``cracking open
the black box,'' as it were \citep{alishahi2019analyzing,blackboxnlp}.
One popular method for studying the linguistic content of neural networks is \defn{probing}, which we define in this work as training a supervised classifier (known as a \defn{probe}) on top of pretrained models' frozen representations \cite{alain2016understanding}. By analyzing the classifier's performance, one can assess how much `knowledge' the representations contain about language.

\begin{figure}
    \centering
    \includegraphics[width=\columnwidth]{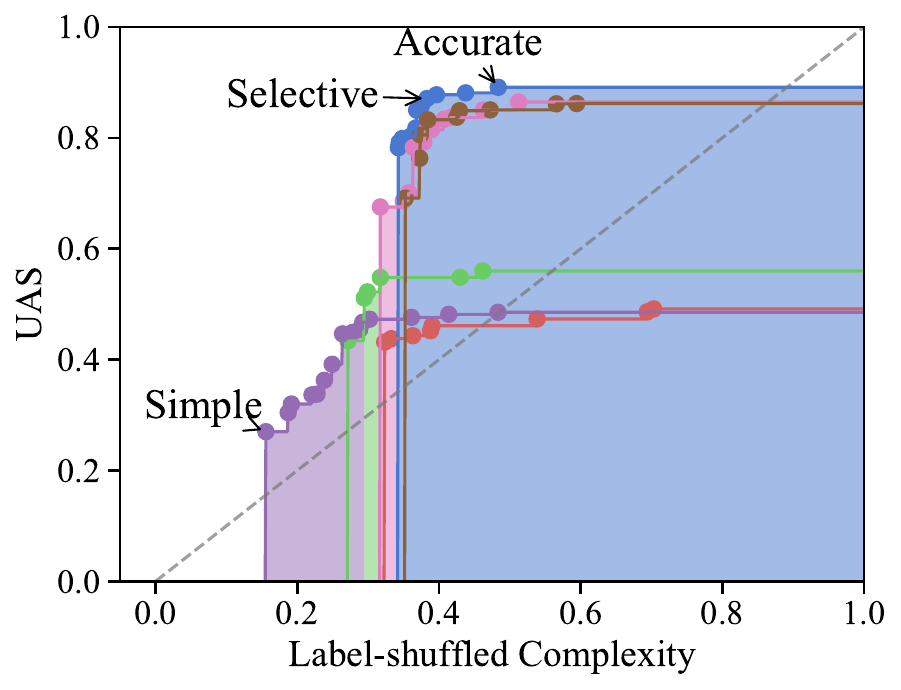}
    \caption{Probe results on dependency parsing in English. The $x$-axis corresponds to complexity and measures a probe's ability to memorize the training data. The $y$-axis measures the probes performance on the task. Probing the representations: \legendalbert, \legendbert, \legendroberta, \legendfasttext, \legendonehot, and \legendrandom.}
    \label{fig:parse_mlp-english}
\end{figure}

Much work in probing advocates for the need for simple probes \cite{hewitt2019structural,maudslay-etal-2020-tale}.
Indeed, on this point, \newcite{alain2016understanding} write: 
\begin{displayquote}
``The task of a deep neural network classifier is to come up with a representation for the final layer that can be easily fed to a linear classifier (i.e., the most elementary form of useful classifier).''
\end{displayquote}
as a justification for their operationalization of \defn{complexity} as the restriction of the probe to linear models (as opposed to deep neural networks).
Most saliently, \newcite{hewitt-liang-2019-designing} attempts to operationalize complexity in terms of control tasks, which constrain a probe's capacity for memorization.\footnote{\newcite{hewitt-liang-2019-designing} define \defn{selectivity} as the difference between a model's accuracy on a task versus its accuracy on a control version of that task. The control version of the tasks are built by randomly shuffling labels across word types and measures a probe's capacity for memorization. Our non-parametric measures of complexity differ from control tasks; we describe these differences in \cref{sec:non-parametric}.}
\newcite{voita2020information} follow in this vein with an information-theoretic estimate of complexity: a model's minimum description length.

In opposition to the complexity of a probe is its \defn{accuracy}, i.e., its ability to perform the target probing task.
From an information-theoretic perspective, \newcite{pimentel-etal-2020-information} argues for the use of more complex probes, since they better estimate the amount of mutual information between a representation and the target linguistic property.
From a different perspective, \newcite{saphra-lopez-2019-understanding} also criticize the indiscriminate use of simple probes, because most neural representations are not estimated with the explicit aim of making information linearly separable; thus, it is unlikely that they will \emph{naturally} do so, and foolish, perhaps, to expect them to. 
\looseness=-1

This paper proposes to directly acknowledge the existence of a trade-off between the two when considering the development of probes. %
We argue---in part based on experimental evidence---that na{\"i}vely selecting a family of probes either for its complexity or its performance leads to degenerate edge-cases; see \cref{fig:parse_mlp-english}.
We conclude that the nuanced trade-off between accuracy and complexity in probing should thus be treated as a bi-objective optimization problem: One objective encourages low complexity and another encourages high accuracy.
We then propose a novel evaluation paradigm for probes. We advocate for \defn{Pareto optimal probes}, i.e., probes that are both simpler and more accurate than all others.
The set of such optimal probes can then be taken in aggregate to form a \defn{Pareto frontier}, which allows for broader analysis and easier comparison between representations.

We run a battery of probing experiments for part-of-speech labeling and dependency-arc labeling, using both parametric and non-parametric complexity metrics.
Our experiments show that if we desire simple probes, then we are forced to conclude that one-hot encoding representations and randomly generated ones almost always encode more linguistic structure than those representations derived from \bert---a nonsensical result. 
On the other hand, seeking the most accurate probes is equivalent to performing NLP task-based research (e.g., part-of-speech tagging) in the classic way. 
We contend our Pareto curve--based measurements strike a reasonable balance. 

To wrap up our paper, we levy a criticism at the probing tasks themselves; we argue that ``toyish'' probing tasks are not very useful for revealing how much more linguistic information \bert manages to capture than standard baseline representations. 
With this in mind, we advocate for more challenging probing tasks, e.g., dependency \emph{parsing} instead of its toyish cousin dependency \emph{arc labeling}.
We find that using actual NLP tasks as probing tasks reveals much more about the advantages \bert provides over non-contextual representations.

\section{Performance and Complexity}\label{sec:trade-off}
We argue in favor of treating probing for linguistic structure in neural representations as a two part optimization problem. 
On the one hand, we must optimize our probe
for \emph{high accuracy} on our chosen probing task: 
If we do not directly train the probe to accurately extract the linguistic features from the representation, how else can we determine whether they are implicitly encoded? 
On the other hand, the received wisdom in the probing community is that probes should be simple \cite{alain2016understanding,hewitt2019structural}:
If the probe is an overly \emph{complex} model, we might ascribe high accuracy on the probing task to the probe itself, meaning the probe has ``learned the task'' to a large extent. 
In this section, we argue that a probing framework that does not \emph{explicitly} take into account the accuracy--complexity trade-off may be easily gamed.
Indeed, we demonstrate how to game both accuracy and complexity respectively below.

\subsection{The Nature of Probing Tasks}\label{sec:toy}
Most \defn{probing tasks} are relatively ``toy'' in nature \cite{hupkes2018visualisation}.\footnote{Not all though, several people have looked into e.g., parse tree reconstruction tasks \cite{jawahar-etal-2019-bert,hewitt2019structural,vilares2020parsing}}
For instance, two of the most common probing tasks
are \defn{part-of-speech labeling} \citep[POSL;][]{hewitt-liang-2019-designing,belinkov-etal-2017-evaluating} and \defn{dependency arc labeling} \citep[DAL;][]{tenney-etal-2019-bert,tenney2019what,voita2020information}.
Both tasks are treated as multi-way classification problems. POSL requires a model to assign a part-of-speech tag to a word in context \emph{without} modeling the entire sequence of part-of-speech tags.
Likewise, DAL requires a model to assign a dependency-arc label to an arc independently of the larger dependency tree. 
These word-oriented probing approaches force models to rely on information about context indirectly encoded in the feature vectors generated by the probed model.
Importantly, both are simplified versions of their structured prediction cousins---part-of-speech tagging and dependency parsing---which require the modeling of entire sentences.
Accuracy on POSL and DAL is then considered indicative of probed representations' ``knowledge'' of the linguistic structure encoded in the probing task.
Limiting \emph{explicit} access to context therefore allows an analysis constrained to how context is \emph{implicitly} encoded in a particular representation.
Furthermore, because POSL and DAL do not require complex structured prediction models, their simplicity is seen as 
a virtue to the mindset of disfavoring complexity (discussed further in \cref{sec:simp}).

\subsection{Optimizing for Performance}\label{sec:acc}
We first will argue that it is problematic to judge a probe either only by its performance on the probing task or by its complexity. 
\newcite{pimentel-etal-2020-information} showed that, under a weak assumption, any contextualized representation contains as much information about a linguistic task as the original sentence. They write:
\begin{displayquote}
``under our operationalization, the endeavour of finding syntax in contextualized embeddings sentences is nonsensical.
This is because, under Assumption 1, we know the answer \textit{a priori}.''
\end{displayquote}
We agree that under their operationalization probing is nonsensical---purely optimizing for performance does not tell us anything about the representations, but only about the sentence itself.

Researchers, of course, have realized that choosing the most accurate probe is not wise for analysis; see \newcite{hewitt2019structural} and the references therein for a good articulation of this point.
To compensate for this tension, researchers have imposed explicit restrictions on the complexity of the probe, resulting in wider differences between contextual and non-contextual representations.
Indeed, this is the logic behind the study of \newcite{hewitt-liang-2019-designing} who argue that \defn{selective}
probes should be chosen to judge whether the target linguistic property is 
well encoded in the representations.
Relatedly, other researchers have explicitly focused on linear classifiers as probes with the explicit reasoning that linear models are simpler than non-linear ones \cite{alain2016understanding,hewitt2019structural,maudslay-etal-2020-tale}.

\subsection{Reducing a Probe's Complexity}\label{sec:simp}
In \cref{sec:acc}, we argued that solely optimizing for accuracy does not lead to a reasonable probing framework. 
Less commonly discussed, however, is that we also cannot directly optimize for simplicity. 
Let us consider the POSL probing task and the case where we are using a linear model as our probabilistic probe:
\begin{equation}\label{eq:linear-probe}
    p(t \mid \vh) = \softmax\left(\vW\vh\right)
\end{equation}
where $t \in \calT$ is the target, e.g., a universal part-of-speech tag \cite{petrov-etal-2012-universal}, $\vh \in \R^d$ is a contextual embedding and $\vW \in \R^{|\calT| \times d}$ is a linear projection matrix. 

A natural measure of probe complexity in this framework is the \defn{rank} of the projection matrix: $\rank(\vW)$. 
Indeed, this complexity metric was considered in one of the experiments in \newcite{hewitt2019structural} to show that \bert representations strictly dominate \elmo representations for all ranks in their analyzed task.
That experiment, though, left out some important baselines---the simplest of which is the encoding of words as one-hot representations. 
We take inspiration from those experiments and expand upon them (but
rely instead on the nuclear norm as a convex relaxation of the matrix rank \cref{sec:parametric}) to produce the more complete plots in \cref{fig:full_english,fig:parameteric}. %
These results are quite stunning; they show that, if we only cared about representations that simple probes could extract linguistic properties from, then a one-hot encoding of the word types is the best choice.

It is easy to see why the one-hot encoding does so well. 
For many of the toy probing tasks, the identity of the word is the single most important factor. 
It seems natural to expect that a low-complexity probe will be unable to exploit much more than a word's identity, so a one-hot embedding is really the best you can do---the word's identity is trivially encoded. 
Our point here is that both accuracy and complexity matter and neither can be sensibly optimized without the other.

\section{An Invitation to Pareto Probing}
We now advocate for a probing evaluation metric that
combines both accuracy \emph{and} complexity.
We argued in \cref{sec:trade-off} that probe accuracy and complexity exist in a trade-off.
Because of this trade-off, we should search for models that are Pareto optimal. 
A probe is considered \defn{Pareto optimal} (with respect to a family of probes) if there is no competing probe where both the accuracy is higher \emph{and} the complexity is lower on the task.
The set of Pareto optimal points may be called the \defn{Pareto frontier} and is generally connected, as is shown in \cref{fig:full_english}. %
As can also be seen in \cref{fig:full_english}, we can compare different representations according to their Pareto frontiers. 
The set of representations that appear on the Pareto frontier should be sufficient--the other representations are \defn{Pareto dominated}, since you can improve in one aspect (complexity or accuracy) without sacrificing the other.
We call the set of representations which are on the frontier \defn{Pareto dominant}.
\looseness=-1

We can also analyze each representation's frontier individually.
This notion leads us to a very natural metric for evaluating probes: \defn{Pareto hypervolume} \cite[PH;][]{auger2012hypervolume}.\footnote{We note that we do not endorse only presenting PH scores, though, since it would again reduces this analysis to a single number. Such scores should be presented together with their Pareto curves to be maximally illustrative.} 
One important technical caveat involving evaluating the hypervolume is that it is undefined when the  metric of model complexity for the experiment is unbounded.
Thus, it is necessary to restrict model complexity to a bounded interval so that the PH is always finite.

\section{Parametric Metrics of Complexity}\label{sec:parametric}
We consider two types of probe complexity metrics.
We term the first \defn{parametric complexity}, which we discuss in this section.
The second type is \defn{non-parametric complexity}, which we discuss in \cref{sec:non-parametric}.
For the parametric one, we first require a family of probes, e.g., the family of linear
probes---which are all those that take the form of \cref{eq:linear-probe}, without restriction on the representation's dimension $d$.

\subsection{Parametric Complexity for Linear Probes}
In the case of linear probes, we explore two metrics of parametric complexity: the nuclear norm and rank. 
The \defn{nuclear norm} is defined as 
\begin{equation}
    ||\vW||_* = \sum_{i=1}^{\min(|\calT|, d)} \sigma_i(\vW)
\end{equation}
where $\sigma_i(\vW)$ is the $i^\text{th}$ singular value of $\vW$---which, in a way, measures the ``size'' of the matrix.
This yields the following objective for $\lambda \ge 0$:
\begin{equation} \label{eq:linear_constraint}
    \underbrace{-\sum_{i=1}^n \log p(t^{(i)} \mid \vh^{(i)})}_{\text{cross-entropy}} + \lambda \cdot \underbrace{\hspace{-.5cm}\phantom{\sum_{i=1}^n} ||\vW||_*}_{\text{nuclear norm}}%
\end{equation}
Training a probe to minimize this objective is equivalent to trading off its performance (high likelihood on the training data) for a lower complexity (nuclear norm of $\vW$).
This trade-off can be controlled through the hyper-parameter $\lambda$.

As a parametric complexity metric, we also consider the \defn{rank} of the matrix. 
One definition of a matrix's rank is the number of non-zero singular values $\sigma_i(\vW)$. 
The rank can easily be restricted to a maximum value $r \in \mathbb{N}_+$ by splitting the matrix in two $\vW = \vW_l^{\top} \vW_r$, where $\vW_l \in \R^{r \times |\calT|}$ and $\vW_r \in \R^{r \times d}$.
The nuclear norm is the tightest convex relaxation of the rank \cite{recht2010guaranteed}.\footnote{Note that one may want to augment the linear probe with a padded vector, i.e.,$\tilde{\vh} = [\vh; 1]$ to include a bias term in the model. In this case, our probe takes the form of $p(t \mid \vh ) = \softmax(\vW\,\tilde{\vh})$ where, now, $\vW \in \R^{|\calT| \times (d+1)}$.}

While low-rank regularization is assumed to produce models that generalize better~\citep{hinton1993keeping,DBLP:journals/corr/abs-1901-10371}, contrary to the classic bias--variance tradeoff, \newcite{goldblum_truth_2019} found that biasing towards small nuclear norms instead \emph{hurts} generalization. 
Furthermore, our probe family consists of linear transformations, which are fed a relatively small number of features and trained with large training sets. 
As such, we are in an underfitting situation and any regularization should indeed hurt test performance.

\subsection{Relation to Minimum Description Length}
A recent proposal by \newcite{voita2020information} suggests
that minimum description length~\citep[MDL; ][]{rissanen1978modeling}
is a useful approach to the problem of balancing performance and complexity. 
The idea behind MDL is analogous to that of Bayesian evidence: We have a family of probabilistic models and a prior over those models. 
The likelihood term tells us how well we have coded the data
and the prior term tells us the length of the model's code:
\begin{equation}
    \underbrace{\prod_{i=1}^n p(t^{(i)} \mid \vh^{(i)}, \vW)}_{\mathrm{likelihood}} \times \underbrace{\hspace{-.25cm}\phantom{\prod_{i=1}^n}p(\vW)}_{\mathrm{prior}}
\end{equation}
If we define our distribution over matrices as 
\begin{equation}\label{eq:normal}
    p(\vW) \propto \exp\left(-\sfrac{\lambda}{2} \cdot ||\vW||^2_*\right)
\end{equation}
we recover our nuclear norm complexity term as the $\log$ of the prior. 
The distribution defined in \cref{eq:normal} is mathematically equivalent to the matrix normal distribution. 
To show this, we note that $-\frac{\lambda}{2}||\vW||_*^2 = -\frac{1}{2}\tr\left(\vW^{\top} \left(\lambda^{-1}\mathbf{I}\right)^{-1} \vW\right)$ and present the definition of the matrix normal \cite[Chapter 2]{gupta2018matrix} as %
\begin{align}
    p(&\vW \mid \textbf{M}, \textbf{V}, \textbf{U}) = \\
    & \frac{\exp\left(-\frac{1}{2} \tr[ \vV^{-1} (\vW - \vM)^\top \vU^{-1} (\vW - \vM) ]\right)}{2 \pi^{kd/2} \vV^{k/2} \vU^{d/2} } \nonumber
\end{align}
where $k=|\calT|$ and $d$ are the sizes of matrix $\vW$, $\vM$ is the matrix's expected value, and $\vV$ and $\vU$ are analogous to the covariance matrices of typical Gaussian distributions.
By setting the mean to the zero matrix, $\vV$ to $\mathbf{I}$ (the identity matrix) 
and $\vU$ to  $\lambda^{-1}\mathbf{I}$, we recover \cref{eq:normal}.
\looseness=-1

Naturally, there are many extensions within the MDL framework, e.g., variational coding MDL \cite{blier_description_2018}.
In the case of non-linear models, Bayesian neural networks \cite{neal2012bayesian} are a natural choice. 
However, a fundamental problem will always remain---the results are dependent on the choice of prior. 
Indeed, in the simple case of linear probes, we can always ``hack'' the prior to favor certain probes over others that may not correspond to our intuitions of model complexity. 
For this reason, we also analyze a set of non-parametric metrics of complexity that do not require the probe user to pre-specify a prior over models.

\section{Non-Parametric Metrics of Complexity}\label{sec:non-parametric}
The parametric metrics of model
complexity in \cref{sec:parametric} have 
an explicit constraint that the models must belong
to the same parametric family. 
Specifically, it requires that we are able to define a penalty (generally dependent on the parameters) that enforces 
how complex each model should be. 
In this section, we move away from parametric notions
of model complexity to non-parametric metrics.

We opt to work with a notion of non-parametric complexity based on the ease with which a model can memorize
training data. 
These non-parametric measures are rarely explicitly discussed as complexity metrics---although they are intuitive for that purpose---and have become common recently: \newcite{zhang_understanding_2016} originated
this trend by shuffling outputs of image data so
the images were no longer predictive of the labels,
using this result to illustrate the effective memorization capacity of modern neural networks. 
The first of our two experiments to obtain non-parametric complexity measures is similar to theirs. We train our probe in a dataset with shuffled labels and get its accuracy in this training set.
We will refer to this complexity metric as the \defn{label-shuffled} scenario.%
\footnote{\newcite{hewitt-liang-2019-designing} use similar methods to create control tasks for their probing experiments. 
Our use of shuffled labels is different from \citeposs{hewitt-liang-2019-designing} in two
important aspects: While they shuffled labels at
the type level, we shuffle them at the token level.
Furthermore, since we are evaluating a model’s capacity for memorization, we look at its accuracy on \emph{the training set}, whereas they
consider the accuracy on the \emph{test set}.}

\begin{figure*}
    \centering
    \includegraphics[width=\textwidth]{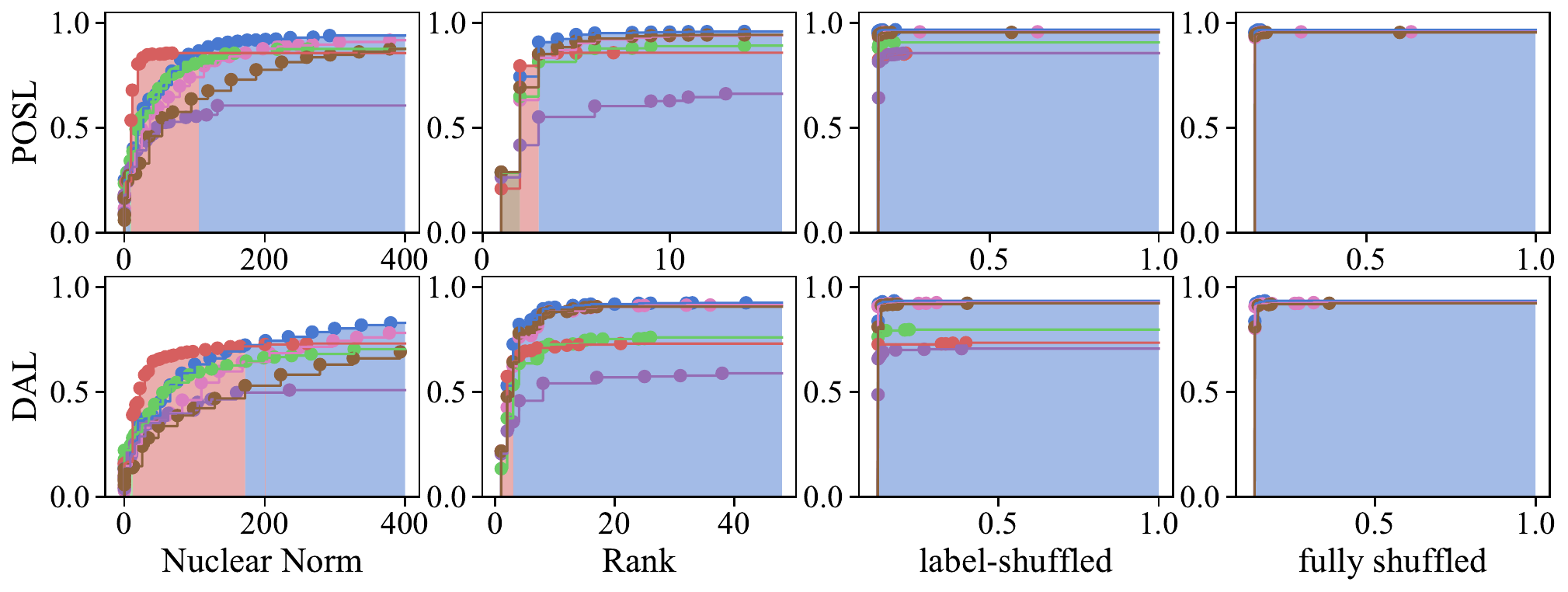}
    \caption{Pareto curves for experiments in English, using both parametric and non-parametric complexity metrics.
    The $x$-axis corresponds to the probe's complexity. The $y$-axis measures the probes accuracy on the task. Probing the representations: \legendalbert, \legendbert, \legendroberta, \legendfasttext, \legendonehot, and \legendrandom.
    }
    \vspace*{-5pt}
    \label{fig:full_english}
\end{figure*}

Neural networks can take advantage of structured input (e.g., real images as opposed to noisy ones) to easily memorize their labels~\citep{zhang_understanding_2016}.
These structured inputs may be easier to represent internally regardless of the outputs, given current theories that early stages of training are committed to memorizing inputs~\cite{arpit2017closer}. 
As such, we may also want to analyze a probe's capacity to memorize unstructured input---%
in the case of language, we can easily remove structure by shuffling the word sequences themselves, creating random Zipfian-distributed noise, which are harder for neural networks to exploit~\citep{liu_lstms_2018}.
By providing probes with unstructured input, we measure a more domain-independent sense of complexity than the ability to map structured inputs to random labels, because the model cannot rely on syntactic patterns when memorizing shuffled training data.
We will refer to this second scenario, wherein both labels and inputs are shuffled, as \defn{fully shuffled}.

The distinction between memorization of real data and memorization of unstructured data is crucial, as experimenters choose the class of probes being learned. 
A comparison between label-shuffled and fully shuffled compression exposes the degree to which the class of probes employs a bias towards the true input structure in its compression.
Similarly, comparisons between different classes of probes can test the same assumed bias. 

We highlight that, while our non-parametric complexity metrics permit arbitrary classes of probes to be included in a probe hypothesis space, the selection criteria of possible probes may still reflect the assumed structure of the data, affecting compression and generalization. 
For example, linear probes reflect an assumption that the information lies in an Euclidean (sub)space;
however, this assumption may not be true: \newcite{reif_visualizing_2019} reveal that a word's sequential position often rests on a spiral manifold in BERT, while the syntactic distances described by \citet{hewitt2019structural} are Pythagorean in nature.
One advantage of these methods is the ability to compare between probe classes, which offers a test of the geometric assumptions behind model selection.\footnote{Another non-parametric method, online coding MDL~\citep{voita2020information} can likewise be compared across arbitrary model classes, because its complexity metric is based on probabilities produced and not probe parameters.} 
Another advantage is that, unlike regularization-based parametric methods, they require no modification of the training procedure and can therefore run much faster.
\looseness=-1

\section{POSL and DAL Experiments}
We present our experimental findings on the previously discussed part-of-speech labeling (POSL) and dependency arc labeling (DAL) probing tasks using both our parametric complexity metrics (\cref{sec:parametric}) and the non-parametric ones (\cref{sec:non-parametric}).
For both tasks, we use data from Universal Dependencies Treebanks version 2.5 \cite{ud-2.5} and we probe a set of 5 typologically diverse languages: Basque (\citealt{aranzabe2015automatic}; BDT licensed under CC BY-NC-SA 3.0), English (\citealt{bies2012english,silveira14gold}; EWT licensed under CC BY-SA 4.0), Finnish (\citealt[TDT]{haverinen2014building} licensed under CC BY-SA 4.0), Marathi (\citealt[UF{\'A}L]{ravishankar2017universal} licensed under CC BY-SA 4.0) and Turkish (\citealt[IMST]{sulubacak-etal-2016-universal} licensed under CC BY-NC-SA 3.0).
When investigating POSL, we take the target space $\calT$ to be the set of universal part-of-speech tags for a specific language.
We then train a classifier to predict these POS tags from word representations obtained from the analyzed model (e.g., \bert). 
Similarly, for DAL, the target space $\calT$ is defined as the set of arc dependency labels in the language, but we predict these labels from pairs of representations---the two words composing the arc.
\looseness=-1

\begin{figure*}
    \centering
    \includegraphics[width=\textwidth]{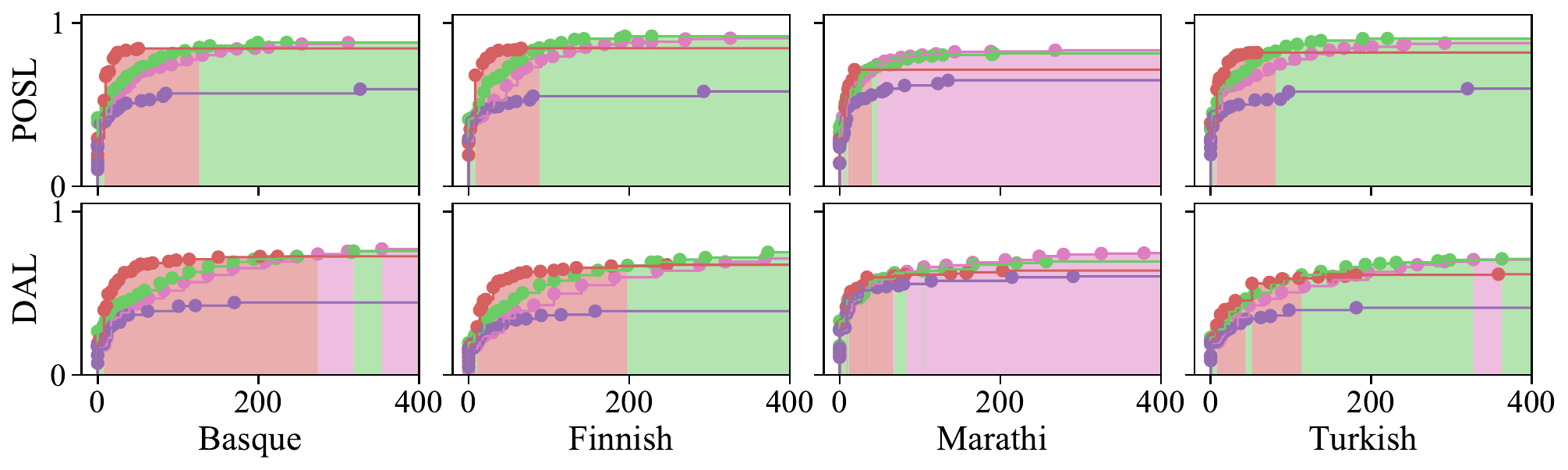}
    \caption{Results for multiple languages using the nuclear norm complexity metric.
    The $x$-axis corresponds to the probe's complexity. The $y$-axis measures the probes accuracy on the task. 
    \legendbert, \legendfasttext, \legendonehot, and \legendrandom.}
    \label{fig:parameteric}
\end{figure*}

We analyze the contextual representations from \bert \cite{devlin-etal-2019-bert}, \albert \cite{lan2019albert} and \roberta \cite{liu2019roberta}---noting that \albert and \roberta are trained in English alone, so we only evaluate their performance on that language.\footnote{We use the base versions (as opposed to the large ones) of \albert, \roberta and multilingual \bert---as implemented by \newcite{wolf2019hugging}}
For each of these models, we feed it a sentence and average the output word piece \citep{wu2016google} representations for each word, as tokenized in the treebank.
We further analyze \fasttext's non-contextual representations \cite{fasttext} as well as one-hot and random representations such as those considered by \newcite{pimentel-etal-2020-information}.
One-hot and random representations map each word type in the training data to a vector we sample from a standard normal distribution (zero mean and unit variance). New representations are sampled on the spot (untrained) for any out of vocabulary words.
All representations are kept fixed during training, except for one-hot, which are learned with the other network parameters.

\subsection{Linear Probes with Norm Constraints} \label{sec:linear_experiment}

For each language--representation--task triple, we train 100 
linear probes, 50 optimizing \cref{eq:linear_constraint}\footnote{We vary $\lambda$ in log-uniform intervals from $2^{-10}$ and $8.0$---while also including experiments with no nuclear norm constraint (i.e., with $\lambda=0$) for completion.} 
and 50 others with the rank constraint.
The left-half of \cref{fig:full_english} presents the Pareto frontiers for both these probes trained on English, while \cref{fig:parameteric} show the nuclear norm experiments in other languages.\footnote{Since rank constrained results showed a similar trend to the nuclear norm ones, results for other languages were moved into the appendices. The interested reader will also find zoomed-in versions (in the $y$-axis) of these plots there, as well as Pareto hypervolume tables.}
As discussed in \cref{sec:simp}, optimizing for complexity alone leads to trivial results---in all these languages one-hot representations would result in the best accuracy when using the nuclear norm complexity metric.
We show that, counter-intuitively, \fasttext and one-hot representations Pareto-dominate \bert on the POSL task in Basque, Finnish and Turkish, producing higher accuracies with probes of any complexity (as defined by their nuclear norms). Thus, from the POSL experiments we cannot conclude \bert has any more syntactic information. 
In English, the one-hot and \albert representations form the Pareto-dominant set; the former in the simple scenario and the later in the complex scenario.
\looseness=-1

\begin{figure}
    \centering
    \includegraphics[width=\columnwidth]{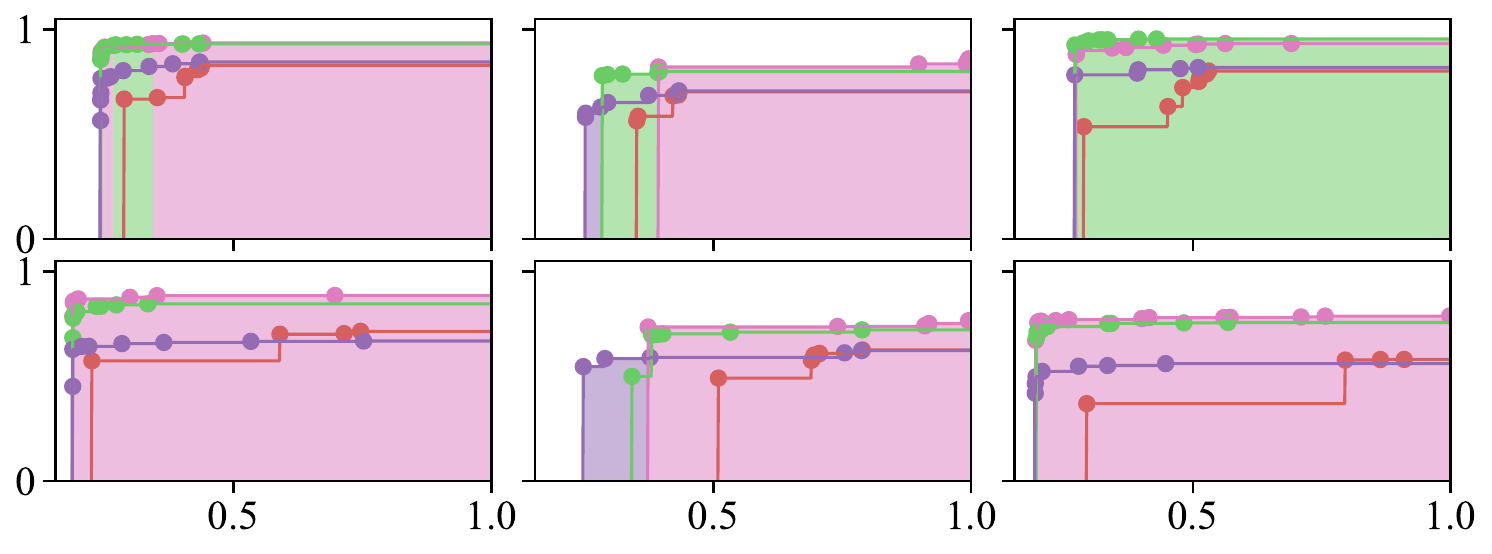}
    \caption{Results using the label-shuffled complexity metric in (left) Basque, (middle) Marathi, and (right) Turkish on (top) POSL, and (bottom) DAL.
    The $x$-axis corresponds to the probe's complexity. The $y$-axis measures the probes accuracy on the task. 
    Probing the representations: \legendbert, \legendfasttext, \legendonehot, and \legendrandom.}
    \label{fig:non-parameteric}
\end{figure}

\subsection{MLPs and Memorization} \label{sec:non-parametric_experiment}

When using our non-parametric complexity metrics, we again train a number of classifiers for each language--representation--task triple. 
The classifiers chosen for this analysis were multilayer perceptrons (MLP) with ReLU non-linearities.
We trained 50 MLPs for each language--representation--task triple, sampling the number of layers uniformly from $[0, 5]$, the dropout from $[0.0, 0.5]$, and the hidden size log-uniformly from $[2^5, 2^{10}]$. Note that zero layers is a linear probe.
(Our MLP is described in \cref{sec:mlp_description}.)
Each of these architectures was trained both on the standard training set as in this set's label-shuffled and fully shuffled alternatives.

\Cref{fig:non-parameteric} presents POSL and DAL multilingual results under the non-parametric complexity metric; the right half of \cref{fig:full_english} presents English results.\footnote{Experiments using the fully shuffled complexity metric only apply to contextual representations, since shuffling the sentence does not affect non-contextual ones. As such we only present them for English---we only analyze one contextual representation in other languages, i.e.,\bert.}
The most striking characteristic of the Pareto frontiers is how simple architectures (i.e.,with relatively low memorization capacity) achieve as high an accuracy as the more complex ones.
This is not surprising, though, when we compare this finding to the parametric ones; there we see  linear probes are already almost as good as MLPs on these tasks.
We take this as support for our intuition that toyish probing tasks are not very interesting or informative. We discuss this point in the next section.

\section{A Call for Harder Probing Tasks}

\subsection{The False Promise of Toy Probing Tasks}
In \cref{sec:toy}, we reviewed arguments that researchers have put forth to justify toy tasks, while the argument for toy tasks from a standpoint of model complexity is addressed in \cref{sec:simp}.
Nevertheless, \bert, \elmo and other pre-trained representations rose to fame based on their ability to boost neural models to human-level scores on large, non-trivial tasks, e.g., natural language inference \cite{liu2019roberta} and question answering \cite{lan2019albert}%
---with different performance patterns being observed on the toyish probing tasks.

\begin{table}
    \centering
    \begin{tabular}{l c c}
    \toprule
Language & POSL & DAL \\
\midrule
Basque & 86\% & 67\% \\
English & 86\% & 68\% \\
Finnish & 87\% & 63\% \\
Marathi & 72\% & 62\% \\
Turkish & 83\% & 58\% \\
    \bottomrule 
    \end{tabular}
    \caption{Test accuracies for a dictionary lookup strategy based on the labels in the training set.}
    \label{tab:results-lookup}
\end{table}

As reported by \newcite{pimentel-etal-2020-information}, \bert embeddings
do not yield substantial improvements over non-contextual-embedding baselines, e.g., \fasttext, on toyish probing tasks.
We reproduce similar experiments, albeit with our methodology, in \cref{sec:non-parametric_experiment}. 
In the case of POSL, we observe that \fasttext's embeddings achieve higher accuracy in many cases. 
In the case of DAL, however, we do observe that \bert leads to relatively small improvements over \fasttext across a typologically diverse set of languages.
This result is not surprising because DAL is a more complex task than POSL: When one probes on simple tasks, models pretrained on more data do not help much.
Furthermore, a quick visual analysis of \cref{sec:linear_experiment} reveals that one can  achieve relatively high accuracy on both POSL and DAL with a linear probe. 
This is confirmed by \cref{sec:non-parametric_experiment}, which shows that simple probes, i.e.,probes with less capacity for memorization, result in as high accuracy as complex ones.
In fact, we run an extra experiment, shown in \cref{tab:results-lookup}, which shows that a trivial dictionary lookup strategy (details are presented in \cref{sec:app_lookup}) already achieves relatively high accuracies in POSL in all languages.

We interpret this to mean that current probing tasks are uninteresting---hiding from us the amount of syntactic information contextual representations actually encode. 
Furthermore, the simplicity of such toyish tasks artificially makes type-level embeddings---e.g., \fasttext---seem nearly as good as contextual ones.
\looseness=-1

\begin{figure}
    \centering
    \includegraphics[width=\columnwidth]{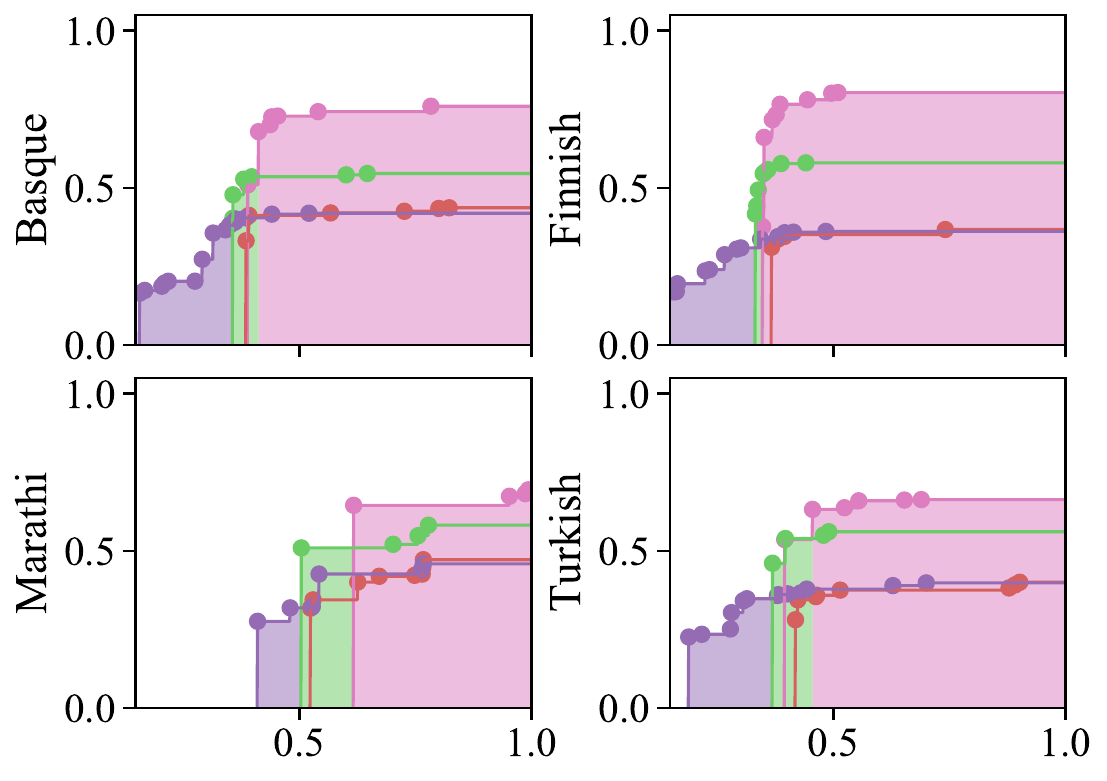}
    \vspace{-15pt}
    \caption{Dependency parsing Pareto curves using the label-shuffled complexity metric in a diverse set of languages.
    The $x$-axis corresponds to the probe's complexity. The $y$-axis measures the probes accuracy on the task. 
    \legendbert, \legendfasttext, \legendonehot, and \legendrandom.}
    \label{fig:non-parametric_parse--full}
\end{figure}

\subsection{Dependency Parsing}\label{sec:dependency}
Following the previous argument, we believe \emph{harder} probing tasks should be used. We take the lead by looking at dependency parsing, which depends on the whole sentence's context, and is much harder than toyish tasks like POSL and DAL.
We train a simplified version of \citeposs{dozat2016deep} biaffine parser, removing its power to process context by discarding its LSTM---as we describe in detail in \cref{sec:parser_architecture}.
This parser gives us the probability of a head for each word in a sentence, which allows us to recover the whole dependency tree.
We then evaluate these trees using unlabeled attachment score (UAS). 
For our label-shuffled experiments, we permute the heads per sentence---creating non-tree dependencies.

\begin{table}
    \centering
    \resizebox{\columnwidth}{!}{%
    \begin{tabular}{l c c c c c}
    \toprule
Representation & Basque & English & Finnish & Marathi & Turkish \\
\midrule
random & 0.32 & 0.39 & 0.28 & 0.24 & 0.30 \\
one-hot & 0.26 & 0.32 & 0.22 & 0.20 & 0.22 \\
\fasttext & 0.35 & 0.40 & 0.38 & \textbf{0.27} & 0.35 \\
\bert & \textbf{0.45} & \textbf{0.57} & \textbf{0.51} & 0.25 & \textbf{0.39} \\
\albert & - & \textbf{0.57} & - & - & - \\
\roberta & - & 0.54 & - & - & - \\
    \bottomrule 
    \end{tabular}
    }
    \caption{Pareto hypervolume results on dependency parsing under the label-shuffled complexity metric.} 
    \vspace{-.5em}
    \label{tab:results-parse}
\end{table}

\Cref{fig:non-parametric_parse--full,fig:parse_mlp-english} present label-shuffled results for this task.
Such figures are much more interesting than the POSL and DAL ones, showing the expected trade-off between accuracy and complexity.
\Cref{tab:results-parse} makes the amount of syntax encoded in contextual representations much clearer when compared to \fasttext. This is specially true if we compare these results to the Pareto hypervolumes of the POSL and DAL tasks (presented in \cref{tab:results-full} in the appendix).
We take this experiment to conclude two things: (i) harder tasks are necessary to study neural representations; (ii) contextual representations encode much more knowledge about syntax (as expected) then do non-contextual ones.

\section{A Closer Look at Model Complexity}

This work represents a new entry into a growing literature on taking the capabilities of probes into account when analyzing a model \citep{hewitt-liang-2019-designing,voita2020information,whitney2020evaluating}.
The fundamental point we wish to espouse in this paper is that
evaluating a probe for linguistic structure is fundamentally asking a question about a trade-off between accuracy and complexity. 
However, we wish to highlight that evaluating a probe's complexity is a very open problem.
Indeed, the larger question of model complexity has been treated for over 50 years in a number of disciplines. 
In statistics, model complexity is researched in the model selection literature, e.g., the classical techniques of \defn{Bayesian information criterion} \citep{schwarz1978estimating} and \defn{Akaike information criterion} \citep{akaike1974new}.
In computer science, learning theorists have introduced
the \defn{Vapnik--Chervonenkis dimension} \citep{vapnik16ja}, \defn{Pollard's pseudo-dimension} \citep{pollard1984convergence}, and \defn{Rademacher complexity} \citep{bartlett2002rademacher}.
Algorithmic information theorists provide \defn{Kolmogorov complexity} \citep{kolmogorov1963tables}---closely related to MDL---encoding the size of the model.
\looseness=-1

A concrete discussion of complexity requires several distinctions regarding these measures. The first is the \defn{object of analysis} of the complexity measure which can be either a \defn{model family}, i.e the whole set of functions realizable by a choice of architecture and hyperparameters, or a \defn{learned model}, which takes into account a specific set of learned parameters. 
The second is the \defn{aspect being analyzed} by the used measure which could be, for example,
the \defn{capacity} of the function class (i.e., whether it is possible to set model weights so the architecture represents a specific target function---its hard constraints) or its \defn{bias} (i.e., the soft constraints that influence whether the training process is likely to guide a model towards this target function).

Importantly, the measure of complexity the scientist employs will impact their scientific findings about how much linguistic structure they read into a neural network's hidden states.
For instance, in some of our experiments we regularize the probe directly with a relaxation of a nuclear norm constraint, thus imposing a bias without modifying the total capacity of the model.
Meanwhile, our rank-based method controls the capacity of the probe directly, enforcing a model family's complexity. 
Finally, our non-parametric methods, and selectivity, estimate the capacity of a model family (under a specific hyperparameter choice) by approximating a ``hard'' function in which labels are randomly assigned---differences in accuracy between these three complexity measures indicate a complex relationship between implicit regularization; input language structure; and model capacity.
In comparison to our \defn{computable} complexity measures, popular \defn{hypothetical} notions also vary in how they explore the data's domain space to analyze a model family: Rademacher complexity uses true observations as inputs, but VC dimension considers adversarially selected data---being defined according to the ``worst'' possible sample allowed in the input domain.
\looseness=-1

As new techniques for considering the complexity of models emerge, it is critical to develop in parallel tools for reasoning about what aspect and object of ``complexity'' is really being measured. When one introduces a metric as modeling complexity, it can be explicitly situated within such a taxonomy; these considerations should be made explicit in the presentation. A sufficiently developed theory of probing will reveal not only the information contained in a representation, but the underlying geometry of the representation space, by comparing the performance of different model families. Such developments are left to future work.

\vspace{-2.5pt}
\section{Conclusion}
\vspace{-1.5pt}

In this paper, we argued for a new approach to probing, treating it as a bi-objective optimization problem.
It has no single optimal solution, but can be analyzed---under the lens of Pareto efficiency---to arrive on a set of optimal solutions.
These Pareto optimal solutions make explicit the trade-off between accuracy and complexity, also permitting for a deeper analysis of the probed representations.%

The second part of our paper argues that we need to select harder tasks for the purpose of probing representations for syntactic knowledge.
For tasks such as POSL or DAL, which require only shallow notions of syntax, non-contextual representations can do almost as well as contextual ones---pretraining on large amounts of data, or encoding contextual knowledge in the representations, does not help much for these tasks.
We then run a battery of experiments on the harder task of dependency parsing; these show that contextual representations indeed provide much more usable syntactic knowledge than non-contextual ones.

\section*{Acknowledgments}
Thanks to Max Balandat for comments on an earlier version of this work.

\bibliography{acl2020}
\bibliographystyle{acl_natbib}

\newpage
\appendix

\section{Multilayer Perceptron} \label{sec:mlp_description}

\newcommand{\relu}{\mathrm{ReLU}}

In this section, we detail the architecture of the MLP classifiers.
We define an MLP block as a series of non-linear transformations:
\begin{align}
    \vh^{(\ell)} = \relu(\vW^{(\ell)}\,\vh^{(\ell-1)})
\end{align}
where $\vh^{(0)} = \vh$ is the initial representation.
The final representation $\vh^{(L)}$ is taken to be the output of an MLP block with $L$ layers.
An MLP block is parameterized by a matrix $\vW^{\ell} \in \R^{d_{\ell} \times d_{\ell-1}}$, where $d_0$ is $\vh$'s dimension, $d_1$ is the MLP's hidden size (as sampled according to \cref{sec:non-parametric_experiment}), and $d_{\ell} = \sfrac{d_{\ell-1}}{2}$ for other layers.
We write: $\vh^{(L)} = \mathrm{MLP}(\vh)$.
The hidden state $\vh^{(L)}$ is then linearly transformed and normalized with a softmax to obtain a probability distribution for either the POSL or DAL tasks:
\begin{align}
    p(t \mid \vh) = \softmax(\vW^{\mathrm{out}}\,\vh^{(L)})
\end{align}

\section{Dependency Parser} \label{sec:parser_architecture}

The dependency parser we make use of is a simplified version of \citeposs{dozat2016deep} biaffine parser. 
The primary simplification is achieved by removing the LSTM, which reduces its complexity as well as restricts its access to context.
We train two MLPs with the same architecture as the one described in \cref{sec:non-parametric_experiment}.
Let $[\vh_1, \ldots, \vh_{|s|}]$ be a sentence representation.
The first MLP is used to model a head dependency representation and the second is used to model a tail dependency relation.
The output of both MLPs is expressed as follows
\begin{align}
    &\vh_{i, \mathrm{head}} = \mathrm{MLP}_{\mathrm{head}}(\vh_i) \\
    &\vh_{i, \mathrm{tail}} = \mathrm{MLP}_{\mathrm{tail}}(\vh_i) \nonumber
\end{align}
which yields a pair of column vectors.
We further define a biaffine function of the two representations as follows
\begin{equation}
    l_{ij} = \vh_{i, \mathrm{head}}^{\top}\, \vW\, \vh_{j, \mathrm{tail}}
\end{equation}
Finally, the output of this biaffine projection is used to construct the probability of  $i$ being the (only) head of $j$, i.e.,
\begin{equation}
    p_{\mathrm{parse}}(i \text{ is the head of } j) =  \frac{e^{l_{i j}}}{\sum_{i'=1}^{|s|} e^{l_{i'j}}}
\end{equation}
Such a parser does not enforce a spanning tree constraint during estimation.
In our experiments, we simply predict each word's head to be the word which maximizes $p_{\mathrm{parse}}(i \text{ is the head of } j)$.
Thus, our predictions will not necessarily be actual trees.
We note, however, that given a trained parser $p_{\mathrm{parse}}$, a dependency tree could be constructed using Edmonds' algorithm.\looseness=-1

\section{Lookup Model} \label{sec:app_lookup}

In this section, we present the design of a very simple lookup model for the POSL and DAL tasks.
We present the detailed implementation of both models below, but their general idea is looking at the training set for an instance's most frequent label, and falling back to an overall general label frequency in case it is not found.

\paragraph{POSL.} In this task, the lookup model has two behaviors: (i) if a word appears in the training set, it guesses its most common label; (ii) if an out-of-vocabulary word, it guesses the most overall frequent label in the training set.

\paragraph{DAL.} In the DAL task, the lookup model has four behaviors: (i) for an arc which appear in the training set, it guesses its most common label; (ii) for an unknown arc, it guesses the most overall frequent label in the training set for the arc's tail word; (iii) if the tail of the arc is an an out of vocabulary word, it guesses the most frequent label in the training set for the head word; (iv) finally, if the head is also out of vocabulary word, it guesses the most overall frequent arc label in the training set.\looseness=-1

\section{Detailed Results}

\begin{figure}[h]
    \centering
    \includegraphics[width=.9\columnwidth]{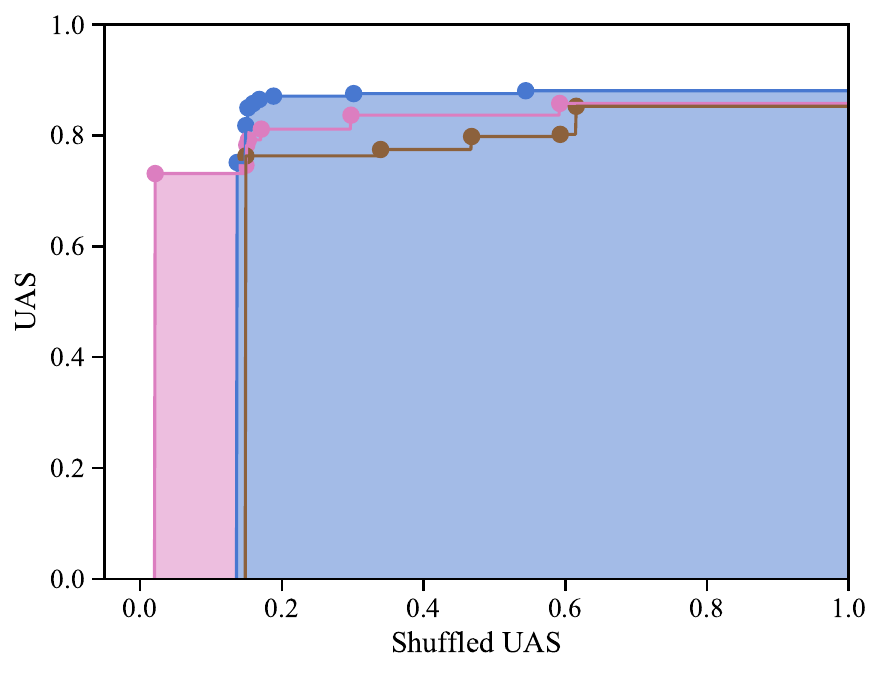}
    \caption{
    Pareto curves on dependency parsing with the fully shuffled complexity metric in English. 
    }
    \label{fig:parameteric_parse-english-fullyshuffled}
\end{figure}

\begin{figure}
    \centering
    \includegraphics[width=\columnwidth]{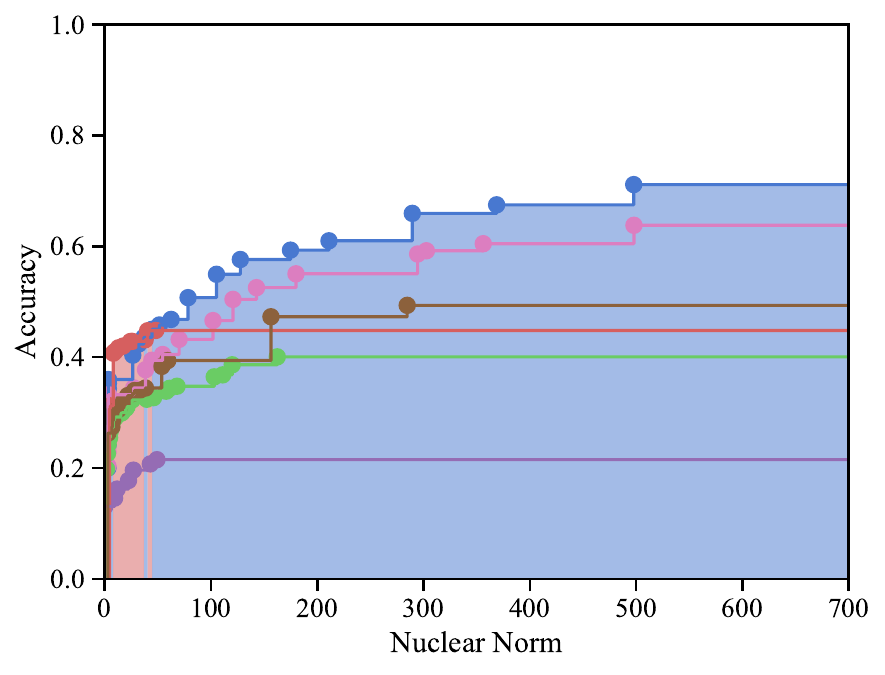}
    \caption{Pareto curves on dependency parsing with the Nuclear Norm complexity metric in English. The $x$-axis corresponds to the complexity, while the $y$-axis measures the probes performance on the task. Since the nuclear norm is unbounded, we maxed it to 700 in the parsing task. Probing the representations: \legendalbert, \legendbert, \legendroberta, \legendfasttext, \legendonehot, and \legendrandom.
    }
    \label{fig:parameteric_parse-english}
\end{figure}

\begin{figure}
    \centering
    \includegraphics[width=\columnwidth]{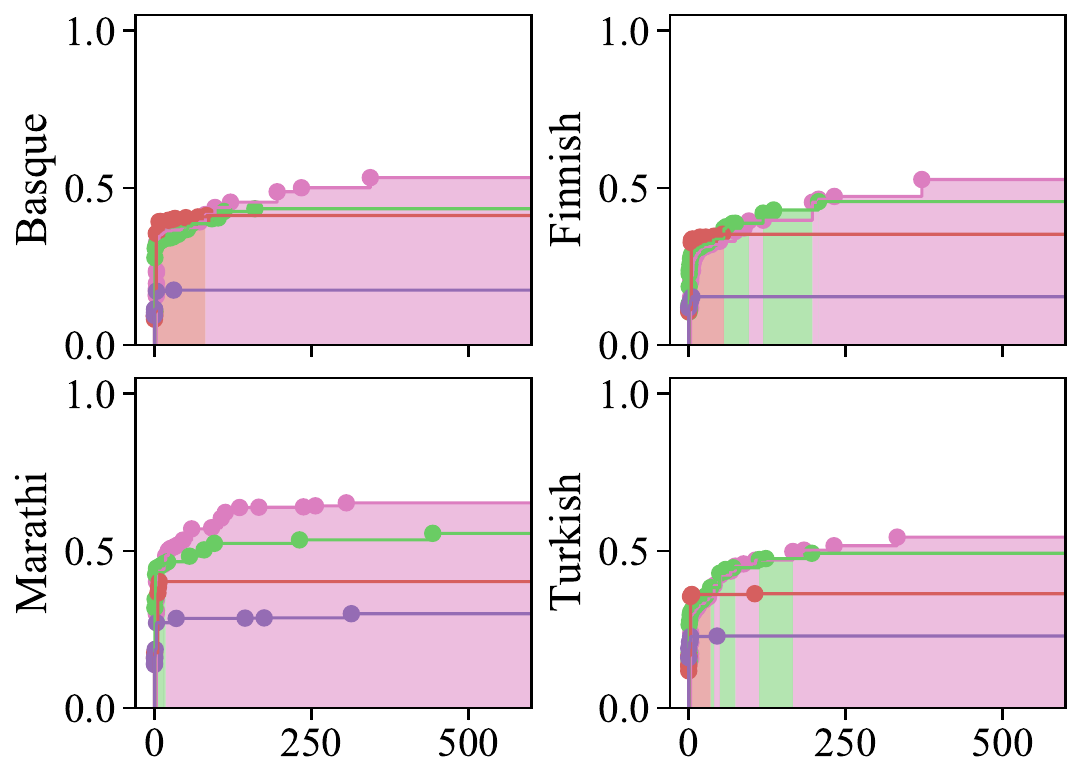}
    \caption{
    Pareto curves on dependency parsing with the Nuclear Norm complexity metric in a diverse set of languages. The $x$-axis corresponds to the complexity, while the $y$-axis measures the probes performance on the task. Since the nuclear norm is unbounded, we maxed it to 700 in the parsing task. Probing the representations: \legendbert, \legendfasttext, \legendonehot, and \legendrandom.}
    \label{fig:parameteric_parse-multi}
\end{figure}

In this section, we present further results which did not fit into the main text.
We initially present results in dependency parsing, using the nuclear norm as our parametric complexity metric.
\Cref{fig:parameteric_parse-english,fig:parameteric_parse-multi} show that, again, the one-hot representation produces the best results in highly constrained scenarios (i.e., with very simple probes).
Furthermore, comparing these results with \cref{sec:dependency} we see that linear probes cannot do as well as MLPs in this task, suggesting it is indeed harder.
\cref{fig:parameteric_parse-english-fullyshuffled} presents fully shuffled results for dependency parsing in English.
Comparing it to \cref{sec:dependency} we see the probes' capacity to memorize is much smaller on unstructured input.

\Cref{tab:results-full} presents the Pareto hypervolume for all the analyzed models, in all languages, for POSL and DAL.
Analyzing this table we again see that contextual representations do not improve over the non-contextual ones on these tasks by much---even producing worse results in some.

\begin{figure*}
    \centering
    \includegraphics[width=\textwidth]{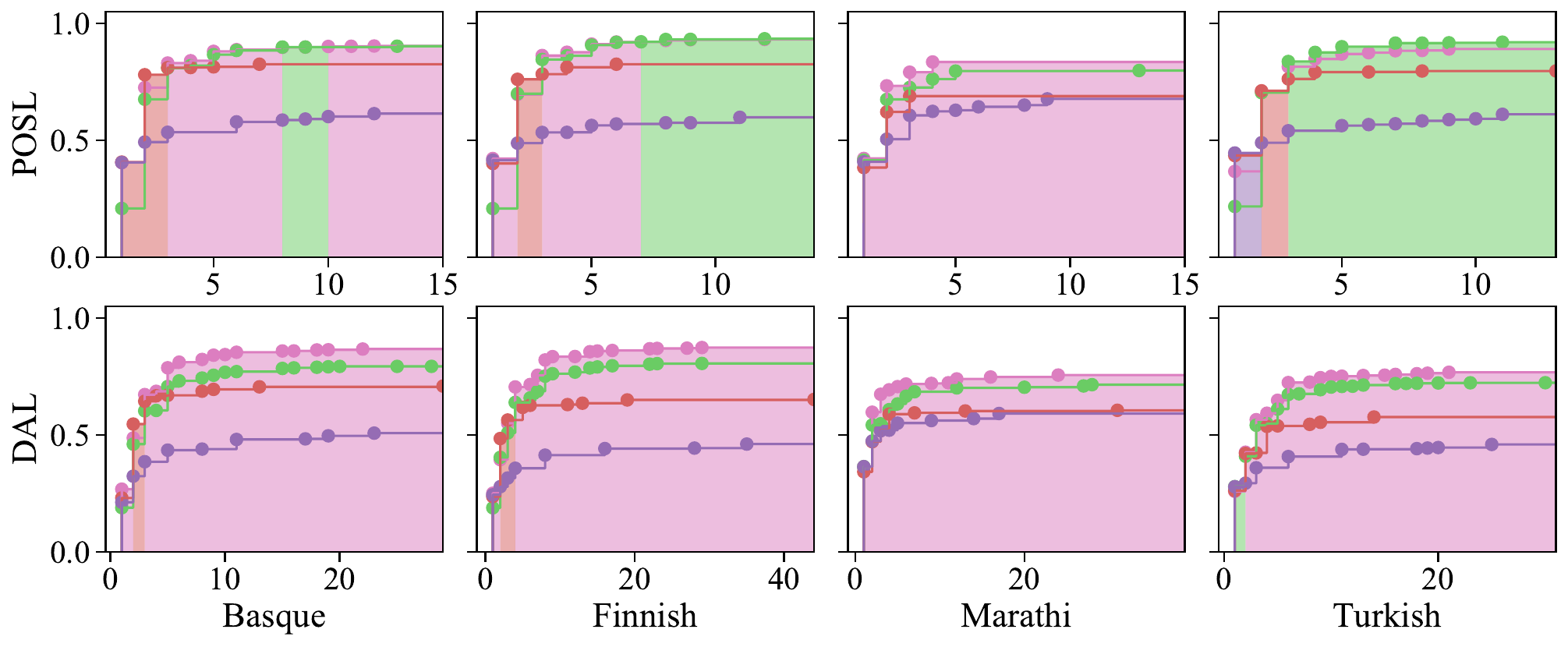}
    \caption{Pareto curves on POSL and DAL with the Rank complexity metric in a diverse set of languages. The $x$-axis corresponds to the complexity, while the $y$-axis measures the probes performance on the task. Probing the representations: \legendbert, \legendfasttext, \legendonehot, and \legendrandom.}
    \label{fig:rank-max}
\end{figure*}

\begin{figure*}
    \centering
    \includegraphics[width=\textwidth]{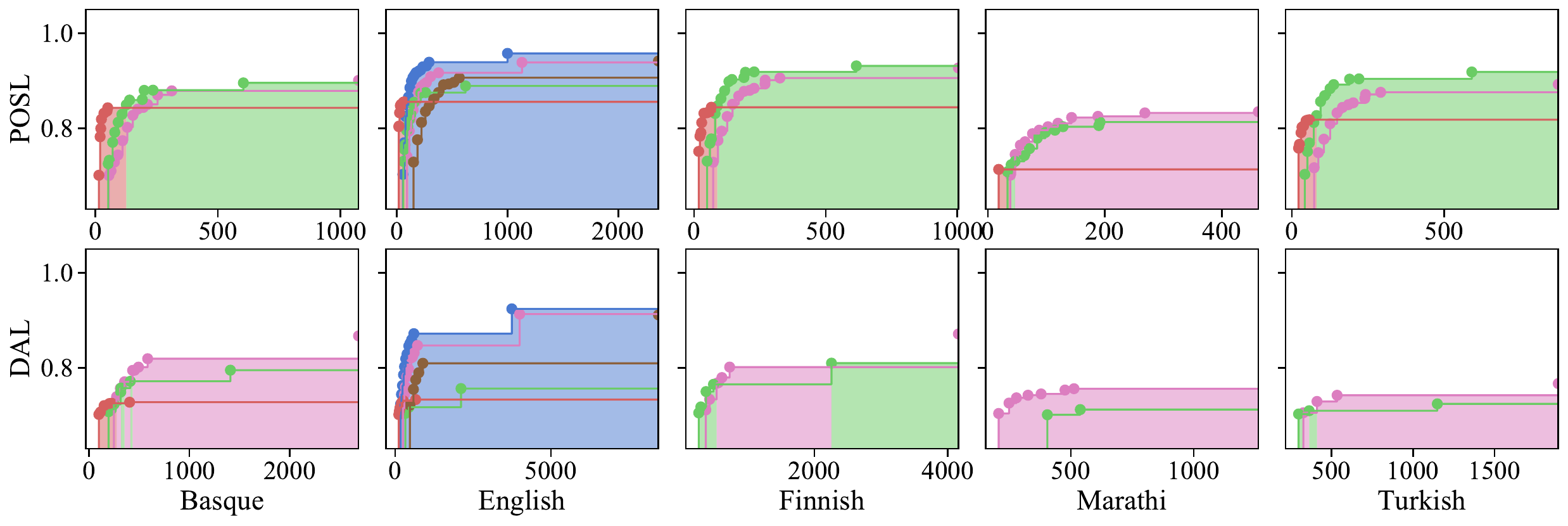}
    \caption{
    Zoomed in (in the $y$-axis) Pareto curves on POSL and DAL under the nuclear norm complexity metric. The $x$-axis corresponds to the complexity, while the $y$-axis measures the probes performance on the task. In this plot we do not max the nuclear norm, showing its full range. Probing the representations: \legendbert, \legendfasttext, \legendonehot, and \legendrandom.}
    \label{fig:parameteric-zoom}
\end{figure*}

Finally, \cref{fig:rank-max} presents results using the Rank parametric complexity metric, while \cref{fig:parameteric-zoom} presents zoomed-in (in the $y$-axis) results for the Nuclear Norm parametric complexity metric.

\begin{table*}[t]
    \centering
    \resizebox{\textwidth}{!}{%
    \begin{tabular}{l l c c c c c c c c c c}
    \toprule
 & & \multicolumn{5}{c}{POSL} & \multicolumn{5}{c}{DAL} \\ \cmidrule(r){3-7} \cmidrule(r){8-12}
Metric & Representation & Basque & English & Finnish & Marathi & Turkish & Basque & English & Finnish & Marathi & Turkish \\
\midrule
Nuclear Norm & \bert & 0.85 & 0.91 & 0.86 & 0.80 & 0.84 & 0.79 & 0.87 & 0.77 & 0.73 & 0.71 \\
Nuclear Norm & \fasttext & 0.86 & 0.87 & 0.90 & 0.78 & 0.88 & 0.76 & 0.74 & 0.77 & 0.69 & 0.70 \\
Nuclear Norm & one-hot & 0.84 & 0.85 & 0.84 & 0.70 & 0.81 & 0.72 & 0.73 & 0.67 & 0.63 & 0.61 \\
Nuclear Norm & random & 0.58 & 0.63 & 0.57 & 0.62 & 0.58 & 0.47 & 0.56 & 0.42 & 0.59 & 0.43 \\
Nuclear Norm & \albert & - & 0.93 & - & - & - & - & 0.89 & - & - & - \\
Nuclear Norm & \roberta & - & 0.87 & - & - & - & - & 0.79 & - & - & - \\
Rank & \bert & 0.78 & 0.80 & 0.80 & 0.74 & 0.75 & 0.78 & 0.85 & 0.80 & 0.71 & 0.69 \\
Rank & \fasttext & 0.76 & 0.77 & 0.78 & 0.70 & 0.76 & 0.71 & 0.70 & 0.74 & 0.66 & 0.65 \\
Rank & one-hot & 0.74 & 0.76 & 0.73 & 0.62 & 0.70 & 0.65 & 0.69 & 0.61 & 0.57 & 0.53 \\
Rank & random & 0.53 & 0.54 & 0.51 & 0.58 & 0.52 & 0.44 & 0.53 & 0.41 & 0.55 & 0.41 \\
Rank & \albert & - & 0.84 & - & - & - & - & 0.87 & - & - & - \\
Rank & \roberta & - & 0.81 & - & - & - & - & 0.85 & - & - & - \\
label-shuffled & \bert & 0.71 & 0.79 & 0.71 & 0.50 & 0.67 & 0.72 & 0.80 & 0.74 & 0.46 & 0.63 \\
label-shuffled & \fasttext & 0.70 & 0.75 & 0.71 & 0.57 & 0.69 & 0.68 & 0.69 & 0.72 & 0.46 & 0.60 \\
label-shuffled & one-hot & 0.57 & 0.70 & 0.59 & 0.45 & 0.52 & 0.50 & 0.64 & 0.53 & 0.28 & 0.30 \\
label-shuffled & random & 0.63 & 0.70 & 0.62 & 0.52 & 0.59 & 0.54 & 0.61 & 0.50 & 0.45 & 0.45 \\
label-shuffled & \albert & - & 0.80 & - & - & - & - & 0.81 & - & - & - \\
label-shuffled & \roberta & - & 0.78 & - & - & - & - & 0.80 & - & - & - \\
fully shuffled & \bert & 0.71 & 0.79 & 0.71 & 0.49 & 0.67 & 0.72 & 0.81 & 0.74 & 0.48 & 0.62 \\
fully shuffled & \albert & - & 0.80 & - & - & - & - & 0.82 & - & - & - \\
fully shuffled & \roberta & - & 0.79 & - & - & - & - & 0.80 & - & - & - \\
    \bottomrule 
    \end{tabular}
    }
    \caption{Pareto hypervolumes for POSL and DAL tasks. Since Nuclear Norm values are unbounded, we limited them to 400. We also normalized Nuclear Norm and Rank results, dividing the volume by their maximum complexity (e.g., the maximum rank or norm).} 
    \label{tab:results-full}
\end{table*}

\end{document}